%% file: naacl.tex
\pdfoutput=1
\documentclass[11pt]{article}

\usepackage[preprint]{acl}

\usepackage{times}
\usepackage{latexsym}

\usepackage[T1]{fontenc}
\usepackage[utf8]{inputenc}

\usepackage{microtype}

\usepackage{inconsolata}

\usepackage{graphicx}
\usepackage{hyperref}
\usepackage{url}
\usepackage{booktabs}
\usepackage{graphicx}
\usepackage{multirow}
\usepackage{wrapfig}
\usepackage{colortbl}
\usepackage{xcolor}
\usepackage{svg}
\usepackage{subcaption} %  for subfigures environments 
\usepackage{algorithm}
\usepackage{algpseudocode}
\usepackage{amsmath}
\usepackage{tikz}
\usetikzlibrary{positioning}
\usetikzlibrary{calc}
\usepackage{tcolorbox}

\definecolor{c1}{cmyk}{0,0.6175,0.8848,0.1490} 
\definecolor{c2}{cmyk}{0.1127,0.6690,0,0.4431} 
\definecolor{c3}{cmyk}{0.3081,0,0.7209,0.3255} 
\definecolor{c4}{cmyk}{0.6765,0.2017,0,0.0667} 
\definecolor{c5}{cmyk}{0,0.8765,0.7099,0.3647} 
\definecolor{forestgreen}{HTML}{397727}

\newcommand{\stanf}{$^\dagger$}
\newcommand{\msr}{$^\diamond$}
\newtcbox{\hlprimarytab}{on line, rounded corners, box align=base, colback=c3!40,colframe=white,size=fbox,arc=3pt, before upper=\strut, top=-2pt, bottom=-4pt, left=-2pt, right=-2pt, boxrule=0pt}
\newtcbox{\hlsecondarytab}{on line, box align=base, colback=red!20,colframe=white,size=fbox,arc=3pt, before upper=\strut, top=-2pt, bottom=-4pt, left=-2pt, right=-2pt, boxrule=0pt}
\newtcbox{\hlgraytab}{on line, rounded corners, box align=base,colframe=white,size=fbox,arc=3pt, before upper=\strut, top=-2pt, bottom=-4pt, left=-2pt, right=-2pt, boxrule=0pt}

\title{Navigating Rifts in Human-LLM Grounding: \\Study and Benchmark}
\author{Omar Shaikh\stanf \thanks{Research performed during an internship at Microsoft.}, Hussein Mozannar\msr, Gagan Bansal\msr, Adam Fourney\msr, Eric Horvitz\msr \\
\stanf Stanford University, \msr Microsoft Research\\
\texttt{oshaikh@stanford.edu}
}

\newcommand{\benchmark}{\textsc{Rifts}}
\newcommand{\copilot}{Bing Chat}

\begin{document}
\maketitle

\begin{abstract}
Language models excel at following instructions but often struggle with the collaborative aspects of conversation that humans naturally employ. This limitation in grounding---the process by which conversation participants establish mutual understanding---can lead to outcomes ranging from frustrated users to serious consequences in high-stakes scenarios. To systematically study grounding challenges in human-LLM interactions, we analyze logs from three human-assistant datasets: WildChat, MultiWOZ, and \copilot{}. We develop a taxonomy of grounding acts and build models to annotate and forecast grounding behavior. Our findings reveal significant differences in human-human and human-LLM grounding: LLMs were three times less likely to initiate clarification and sixteen times less likely to provide follow-up requests than humans. Additionally, we find that early grounding failures predict later interaction breakdowns. Building on these insights, we introduce \benchmark{}, a benchmark derived from publicly available LLM interaction data containing situations where LLMs fail to initiate grounding. We note that current frontier models perform poorly on \benchmark{}, highlighting the need to reconsider how we train and prompt LLMs for human interaction. To this end, we develop a preliminary intervention aimed at mitigating grounding failures.
\end{abstract}

\input{sections/01_introduction}
\input{sections/02_related_work}

\input{sections/03_grounding_acts}

\input{sections/04_data}
\input{sections/05_modeling}

\input{sections/06_benchmark}
\input{sections/07_discussion_conclusion}

\bibliography{naacl}

\newpage
\input{sections/99_appendix}
\end{document}

%% file: sections/01_introduction.tex
\section{Introduction}

\begin{figure}[ht!]
    \centering
    \includegraphics[width=\linewidth]{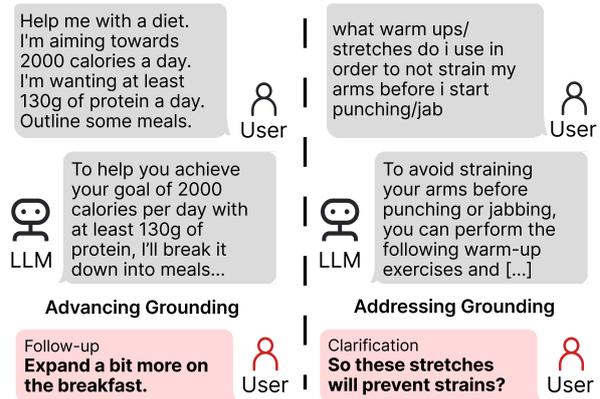}
    \caption{\textbf{People initiate grounding acts more frequently than LLMs.} In settings where a forthcoming turn \textit{advances} grounding (left), people are more likely than LLMs to initiate follow-ups and refine interaction goals. In situations where grounding challenges are \textit{addressed} (right), repairs are primarily initiated by people.}     
    \label{fig:teaser_fig}
\end{figure}

Language models used for conversational interaction are trained primarily to follow instructions~\cite{ouyang2022training}.  But effective dialogue requires more than just instruction-following. Participants in conversation work together in a collaborative process, resolving ambiguities as they exchange ideas and achieve shared objectives. They cultivate \emph{common ground}\footnote{We use \emph{grounding} to refer to Clark's formulation of the language, gestures, and signaling that participants in a conversation employ to establish and maintain an effective dialogue with shared understanding among actors~\cite{clark1996using}.} through \emph{grounding processes}: communicative behaviors aimed at establishing and confirming mutual comprehension~\cite{clark1996using}. Grounding mechanisms guide how individuals anticipate and react to a conversation partner's contributions~\citep{clark1989contributing}. Speakers implement grounding through dialogue acts. They work to confirm or \textit{clarify} assumptions, and ask \textit{follow-up questions}. When grounding breaks down, participants employ \textit{repair} strategies to resolve potential miscommunications.

In contrast, LLMs employed in conversational systems generate task-centric content with minimal grounding actions \cite{shaikh2024grounding}. Failing to ground with users can be costly, with outcomes ranging from the common situation of frustrated users to that of serious consequences in high-stakes situations (Figure~\ref{fig:teaser_fig}). The process by which humans build common ground offers a useful lens through which tounderstand human-computer interaction~\cite{brennan2014grounding}. Ideal human-LLM grounding should allow for both human and machine initiatives, aimed at detecting misunderstandings and achieving mutual understanding~\cite{horvitz1999principles}. 

We start by assessing the current state of grounding in human-LLM interactions, analyzing grounding behavior through real-world interaction logs. To examine human-LLM grounding, we first characterize interactions through a set of validated dialogue acts (\S\ref{sec:grounding_acts}). In synthesizing these grounding acts, we build on prior work in conversational analysis and dialogue systems.  We cover acts that communicate progress (acknowledgment, follow-up, etc.) and difficulty in grounding (repair, clarification, etc.), initiated by either a human or LLM.

Using a set of grounding acts, we construct models of human-LLM grounding, which we use to annotate and to forecastgrounding in conversation (\S\ref{sec:data}-\ref{sec:modeling_all}). We construct and validate an LLM-based \emph{annotator}, which enables efficient annotation of logs from publicly available interactions with ChatGPT (WildChat), data from a widely used commercial LLM service (\copilot), and Wizard-of-Oz-ed interactions with a human roleplaying as an AI (MultiWOZ). We additionally develop a grounding forecaster that predicts the presence of grounding acts in future turns. The forecaster enables a dialogue intervention that can proactively prevent grounding difficulties in forthcoming turns.

Our analysis of human-LLM interaction data reveals significant asymmetries in initiating grounding: people are three times more likely to clarify and sixteen times more likely to issue follow-up requests compared to LLMs. In addition, we find that grounding failures occurring early in dialogue cascade into higher likelihoods of downstream failure. Overall, there is substantial room for improvement in human-LLM interaction via addressing deficits in conversational grounding.
 
To systematically measure human-LLM grounding and test interventions, we introduce a new benchmark (\S\ref{sec:final-eval}). \benchmark{}\footnote{\benchmark{} can be accessed at this link: \href{https://github.com/microsoft/rifts}{https://github.com/microsoft/rifts}.} is a curated set of $\approx$ 1.8K tasks---directly sourced from in-the-wild interaction logs---that require \emph{selective} use of clarification and follow-up requests for interactive grounding. Most frontier models struggle with \benchmark{}. While we propose an effective intervention with our forecaster, progress on \benchmark{} will require rethinking how LLMs are trained to interact with people.

%% file: sections/02_related_work.tex
\section{Related Work}

\paragraph{Ambiguity and Common Ground}
Disambiguating questions like “\textit{Did you see (the man (with the telescope))?}” requires establishing common ground. While NLP benchmarks address ambiguity~\cite{min2020ambigqa, tamkin2022task}, they focus on well-defined cases (e.g., reference ambiguity). However, people engage LLMs for open-ended tasks, e.g., creative writing and cover letters---where correct answers aren’t predefined, making common ground crucial. Using naturalistic human-LLM interactions, we identify challenges in building common ground. To operationalize it~\cite{clark1989contributing, clark1996using, stalnaker2002common}, we synthesize dialogue acts based on subdialogues~\cite{Lit85, LitAllCs87} and conversation structure~\cite{jefferson1972side}.

\paragraph{Grounding in Dialogue Systems} Numerous NLP systems from ELIZA onward have been designed to initiate some form of conversational grounding~\cite{eliza_citation, purver2004theory, li2023eliciting, paranjape2021human}. Decision-theoretic models have helped systems manage uncertainty about user goals~\cite{paekUAI2000conversation, horvitzpaek2000Interspeech,paek2003utility}, while multimodal approaches consider language and visual cues \cite{pejsa2014natural}. Human-LLM grounding is crucial for tasks including goal-coordination \cite{bara2021mindcraft,mohanty2023transforming,fried2022pragmatics,li-boyer-2015-semantic}, planning~\cite{Chu98,Loc98}, games~\cite{madureira2023instruction, shaikh-etal-2023-modeling}, data retrieval~\cite{lu-etal-2023-statcan}, improvisation~\cite{cho-may-2020-grounding}, and design~\cite{vaithilingam2024imagining}. AI-initiated grounding improves conversation quality \cite{zhou2022reflect}, enables human-AI collaboration~\cite{lin2023decisionoriented}, and encourages humans to reflect on LLM outputs~\cite{park2023thinking}. Our work extends this by introducing methods and a benchmark for studying grounding in real-world human-LLM dialogue.

\paragraph{Proactive Mitigation of Grounding Failures}
Research has explored the use of machine-learned models to predict and mitigate grounding failures in spoken dialogue systems. By forecasting potential failures, models can guide proactive interventions. For example, in a call-routing system, predictions of downstream grounding issues were used to trigger early transfers to human operators, reducing user frustration and disengagement, like pressing keys to access human assistants or abandoning calls~\cite{horvitz2007complementary}.

\begin{table*}
    \centering
    \small
    \begin{tabular}{l|ll}
        \toprule
        \textbf{Type} & \textbf{Act} & \textbf{Example}  \\ 
        \midrule
         & \texttt{Instruction} & \includegraphics[width=0.8em]{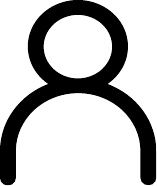} Write a story. \\
        \midrule
        \hlprimarytab{\textbf{Advancing}}
            & \texttt{Next Turn} 
            & \includegraphics[width=0.8em]{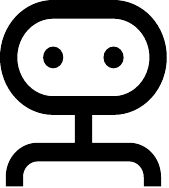} \hlprimarytab{Once upon a time, in a mysterious forest...} 
            \\ 
            & 
            & \multicolumn{1}{l}{\textit{Attempting to advance grounding with the next relevant turn.}} \\
            & 
            & \multicolumn{1}{l}{\textit{Note: all other successful acts are a \textbf{subset} of} \texttt{Next Turn.}} \\
        \cmidrule{2-3}
            & \texttt{Acknowledge} 
            & \includegraphics[width=0.8em]{images/robot.pdf} \hlprimarytab{I understand. [I will write you a story.]} Once upon... 
            \\ 
            & 
            & \multicolumn{1}{l}{\textit{Verbalizing understanding: ``I see, O.K.'' / repeating instruction.}} \\ 
        \cmidrule{2-3}
            & \texttt{Follow-up} 
            & \includegraphics[width=0.8em]{images/robot.pdf} Once upon a time... 
             \\
            & 
            & \includegraphics[width=0.8em]{images/human.pdf} \hlprimarytab{Can you make it longer?}
             \\
        \midrule
         \hlgraytab{\textbf{Disambiguating}} & \texttt{Overresponse} 
            & \includegraphics[width=0.8em]{images/robot.pdf} \hlgraytab{Writing a story requires a plan: First .... Also,} here is an example story.... 
            \\ 
        \cmidrule{2-3}
            & \texttt{Clarify} 
            & \includegraphics[width=0.8em]{images/robot.pdf} \hlgraytab{Do you want a story or plan to write one too?} 
            \\ 
        \midrule

         \hlsecondarytab{\textbf{Addressing}}
            & \texttt{Repair} 
            & \includegraphics[width=0.8em]{images/robot.pdf} \texttt{[Overresponse]} 
            \\ 

            & 
            & \includegraphics[width=0.8em]{images/human.pdf} \hlsecondarytab{Just give me the story, nothing else.} 
            \\ 
        \cmidrule{2-3}
            & \texttt{Reformulate} 
            & \includegraphics[width=0.8em]{images/robot.pdf} \texttt{[Next Turn]} - incorrectly assumed common ground. 
            \\
            & 
            & \includegraphics[width=0.8em]{images/human.pdf} \hlsecondarytab{Please write a story.}
            \\
        \cmidrule{2-3}
            & \texttt{Restart} 
            & \includegraphics[width=0.8em]{images/robot.pdf} \texttt{[Next Turn]} - incorrectly assumed common ground. 
            \\
            & 
            & \includegraphics[width=0.8em]{images/human.pdf} \hlsecondarytab{\texttt{[User leaves session and restarts with the same instruction.]}}
            \\
        \bottomrule
    \end{tabular}
    \caption{Examples of actions we formulated for understanding grounding in multi-turn human-LLM interaction. These acts serve as proxies for a participant that attempts to \hlprimarytab{advance} grounding, \hlgraytab{disambiguating}, or \hlsecondarytab{address} grounding.}
    \label{tab:act_definitions}
\end{table*}

\paragraph{LLMs and conversational grounding} Current LLMs appear to guess user intent and progress with assumptions of correct inferences rather than resolving uncertainties through grounding.\footnote{\url{https://openai.com/blog/chatgpt}} This manifests as generations of over-informative responses \cite{tsvilodub2023overinformative}, refusal to handle ambiguity~\cite{abercrombie2023mirages,min2020ambigqa,gao2021answering}, and overconfidence~\cite{mielke2022reducing}. Prior work demonstrated that LLM-powered conversational agents fail to generate appropriate grounding acts~\cite{shaikh2024grounding, lu2024newsinterview}. Rather than measuring human-LLM interaction through end-to-end evaluation \cite{lee2022evaluating,chiang2024chatbot}, we consider discrete grounding acts in dialogues. Like \citet{schneider2024bridging} and \citet{shaikh2024grounding}, we use prompted LLMs to classify these acts. While prior work has explored generating clarification requests through prompting~\cite{kuhn2022clam, chen-etal-2023-controllable} and finetuning~\cite{andukuri2024star, zhang2023clarify, hong2023zero, gan2024clarq}, we examine grounding acts more broadly.

%% file: sections/03_grounding_acts.tex
\section{Human-LLM Grounding Acts}
\label{sec:grounding_acts}
To measure grounding between people and LLMs, we curate a set of dialogue acts that serve as proxies for grounding. We outline our selected acts and discuss prior work motivating each act. Our typology builds on prior work in conversational grounding: ~\citet{clark1989contributing} and \citet{traum1992conversation} outline a hierarchy of actions, including discourse acts, that are used by humans to ground with one another. Recent work has revisited conversational grounding in the context of LLMs~\cite{shaikh2024grounding, schneider2024bridging}, focusing on a subset of acts generated mainly by people (e.g, following-up, acknowledging understanding, and clarifying).

In contrast, we consider acts generated by both LLMs and people. Our selected acts serve as signals for effective grounding; we segment acts across \emph{communicated} grounding outcomes. We focus on \hlprimarytab{advancing} the construction of common ground, \hlsecondarytab{addressing} a potential grounding failure, or \hlgraytab{disambiguating}. Using our typology, we can measure grounding outcomes with observable dialogue acts during human-LLM interaction (illustrative examples in Table \ref{tab:act_definitions}). 

\subsection{\hlsecondarytab{Addressing Acts}}
Addressing acts are made in response to detection of inadequate grounding. They explicitly signal a potential misunderstanding. Here, participants engage with a focus on addressing the failure. 

\emph{Reformulations} occur when a participant repeats or restates their query in other words because of a failure to ground. An utterance is a reformulation if the succeeding utterance from the same participant is semantically equivalent to the original. Reformulations are prevalent in search engine and information retrieval domains ~\cite{lau1999patterns}.

\emph{Repairs} also signal a grounding failure. Unlike reformulations, the listener \textit{directly} corrects a misunderstanding from another speaker (e.g. \textit{I meant do it in JavaScript, not Python.})~\cite{schegloff1992repair, schegloff1977preference}

\emph{Restarts} occur when users reset a conversation to improve understanding and achieve a successful dialogue. They often follow significant misunderstandings, whether in the initial response or across multiple exchanges. Users may restart due to LLMs misinterpreting intent, ambiguity, sensitivity, or irrelevant context~\cite{shi2023large}, akin to search query retries after irrelevant results~\cite{lau1999patterns}. Research on restarts includes user decisions to suspend a dialogue and seek alternate solutions, such as transferring from AI-based systems to human operators before frustration escalates~\cite{horvitz2007complementary}. Restarts are classified when an initial instruction is repaired or reformulated within 30 minutes~\cite{radlinski2005query, downey2007models}.

\subsection{\hlgraytab{Disambiguating Acts}}
The status of grounding may also be uncertain between participants. Rather than clearly indicating success or failure, \emph{disambiguating} acts represent strategies that participants use to---potentially inefficiently---lower the likelihood of potential misunderstandings.

\emph{Clarifications} occur when a participant seeks to disambiguate an utterance from another participant; or when a participant proactively ``clears up'' misunderstandings. Clarifications often occur when the task at hand is perceived as ambiguous (e.g. \textit{What did you mean by that?}).~\cite{ginzburg2001resolving,purver2003answering,purver2003means,healey2011making,healey2003experimenting,purver2004theory,stoyanchev13,madureira2023you,rahmani2023survey}.

\emph{Overresponses} include \textit{more} than what another participant reasonably asked for. Unlike Next Turn, which provides only \textit{expected} information, overresponses also anticipate and respond to potential follow-ups, flouting the Gricean maxim of quantity~\cite{grice1975logic}. Overresponses often appear as overly \textit{verbose} LLM-generated answers---a behavior contemporary reward models are criticized for encouraging by favoring longer responses~\cite{singhal2023long}.

\subsection{\hlgraytab{Advancing Acts}}

Advancing acts signal that a participant \textit{understands} utterances from another participant. 

\emph{Next Turns} refer to the next conversational move made by a listener that is expected, given the prior turn(s) in a dialogue. Examples of relevant next turns include directly answering a question, expressing an opinion (agreeing or disagreeing), or apologizing. If no misunderstanding has occurred, a listener has moved on to the next relevant turn \textit{by default.}~\cite{levinson1983pragmatics, sacks1978simplest, schiffrin1987discourse}. Note that advancing grounding will initiate the Next Relevant Turn. We focus on two: follow-ups \& acknowledgments. 

\emph{Follow-ups} elaborate on a prior utterance in an interaction. Unlike clarifications---which disambiguate or clarify---follow-ups seek additional information. Because follow-ups build on a prior utterance, they implicitly signal understanding of past utterances.~\cite{davis1982determinants, graesser1995collaborative, traum1992conversation, bunt2017dialogue}.

\emph{Acknowledgments} explicitly signal understanding. These requests manifest either through explicit dialogue (e.g. \textit{I see; I understand; O.K.}) or by repeating portions of another participant's utterance (e.g. \textit{I can help you [write a story].}) Unlike reformulation, where a listener repeats to address failure, acknowledgment occurs when a speaker repeats to demonstrate understanding.~\cite{schegloff1982discourse, sacks1978simplest, schiffrin1987discourse, clark1989contributing, cho-may-2020-grounding}

%% file: sections/04_data.tex
\section{Data}
\label{sec:data}
To analyze collaborative grounding with LLM-based assistants, we draw from three English-language datasets consisting of dialogues between a human and an assistant. WildChat is a real-world human-AI interaction dataset with interaction between people and several OpenAI models~\cite{zhao2024wildchat}. User data was collected with consent, in exchange for free access to the models. We use the non-toxic version of WildChat, filter conversations to be in English, and sample one conversation from each user. \copilot{}, similar to WildChat, was collected from a large chat-based service used by millions of users, powered by OpenAI LLMs. Finally, MultiWOZ is a crowdsourced dataset of dialogue-based interaction with an assistant~\cite{budzianowski2018large, ramadan2018large, eric2019multiwoz, zang2020multiwoz}. In contrast to WildChat and \copilot{}, MultiWOZ contains human-human dialogues, with one human playing a ``wizard-of-oz'' role as the assistant. We use MultiWOZ 2.2~\cite{zang2020multiwoz}, examining collaborative grounding acts on the validation and test splits for a subset of the tasks.
While we observe similarities in terms of the tasks posed by humans across the three datasets, differences do exist, which makes direct comparisons difficult. In Appendix \ref{appdx:full_dataset_splits}, we outline the number of dialogues and messages in each dataset.

%% file: sections/05_modeling.tex
\section{Modeling Human-LLM Grounding}
\label{sec:modeling_all}
Given a defined set of grounding acts and data drawn from logs of human-assistant interaction, we can build grounding models. We first build prompted classifiers that identify acts post-hoc and describe the status quo of human-LLM grounding (\S\ref{sec:modeling_classification}). Then, we train grounding forecasters, enabling forecasting of grounding acts given just the initial instruction (\S\ref{sec:forecasting-grounding}). With forecasters, we can identify tasks where grounding is critical and intervene when appropriate.

\subsection{Classifying Human-LLM Grounding Acts}
\label{sec:modeling_classification}
\paragraph{Method.} We employ GPT-4o-mini to annotate grounding acts across a subset of our datasets. On subsets of our data, we observed nearly identical results using GPT-4o compared to GPT-4o-mini. Thus, we employ GPT-4o-mini for its affordability to allow for efficiency and reproducibility. 
Following~\citet{shaikh2024grounding}, we first encode our typology in a prompt (Appendix \ref{appdx:grounding_acts_prompt}) and label each turn in the conversations. To validate the accuracy of the approach, three authors annotated 10 dialogues (total of 108 messages) from each dataset. We found that a great deal of the early disagreement among annotators was in assessing clarification vs. follow-up questions. Disgreements were resolved through a round of discussion and reannotation, reaching an average Cohen Kappa of 0.71 across the datasets. For the final dataset, ties were broken through discussion, converging on a final selection of the majority label. Using this as a withheld test set, we find that Macro F-1 scores are reasonably high across datasets (0.75). A full table of results across labels is also in Appendix~\ref{appdx:grounding_acts_prompt}. 
% \vspace{3in}
\begin{figure}[t]
    \centering
    \includegraphics[width=0.92\linewidth]{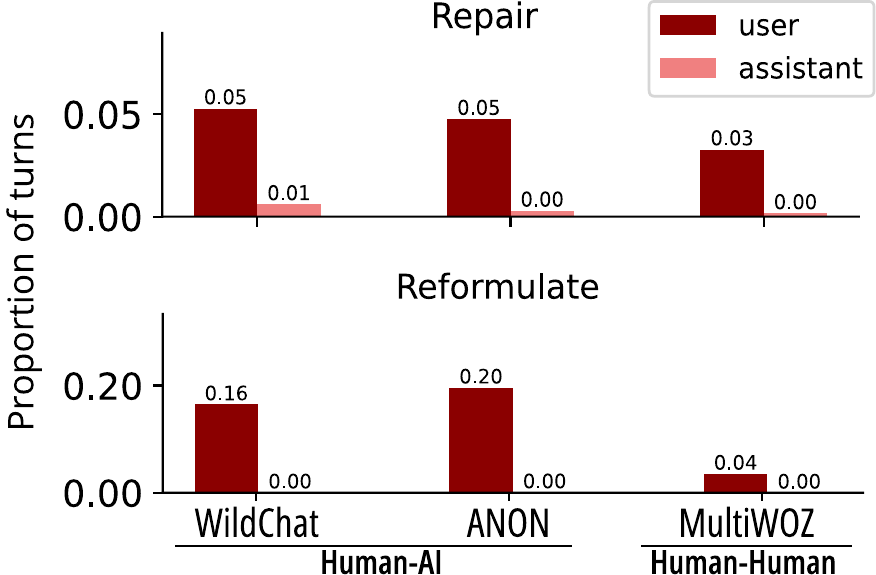}
    \caption{\textbf{Addressing Acts.} In human-LLM interaction---WildChat and \copilot{}---we observed high rates of repair (row 1) and reformulation (row 2) from human users. In contrast, users repair/reformulate less when interacting with a human wizard-of-oz-ing as an assistant (MultiWOZ).}
    \label{fig:neg_grounding}
\end{figure}

\begin{figure}[t]
    \centering
    \includegraphics[width=0.92\linewidth]{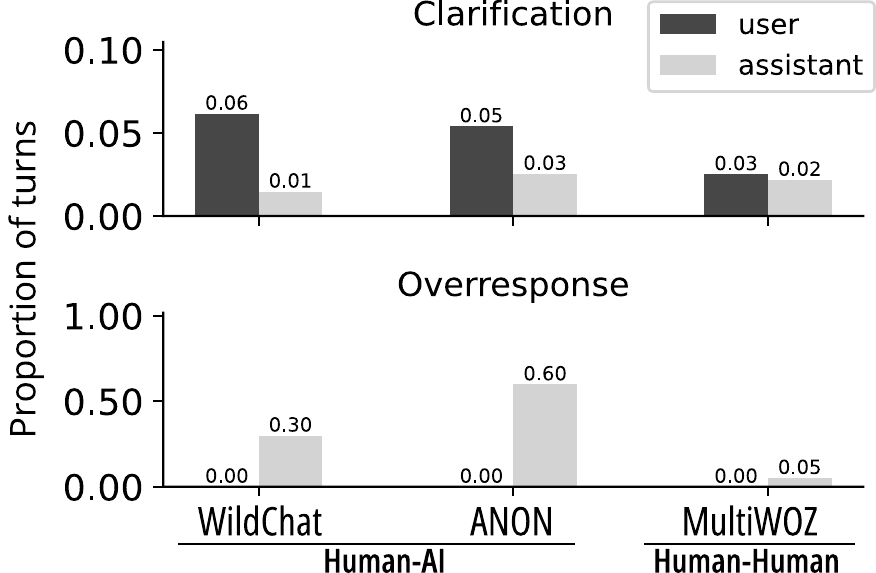}
    \caption{\textbf{Disambiguating Acts.} LLM assistants infrequently initiate {\em clarification} to avoid human repair (row 1). In Human-LLM interaction (WildChat, \copilot{}), \emph{users} clarify at significantly higher rates than assistants. Human-human interaction (MultiWOZ), however, has similar rates of clarification from both users and assistants. Instead of clarifying, LLM assistants regularly overrespond, disambiguating by generating \textit{more} than what the user reasonably asked for (row 2).}
    \label{fig:ambig_grounding}
\end{figure}

\begin{figure}[t]
    \centering
    \includegraphics[width=0.92\linewidth]{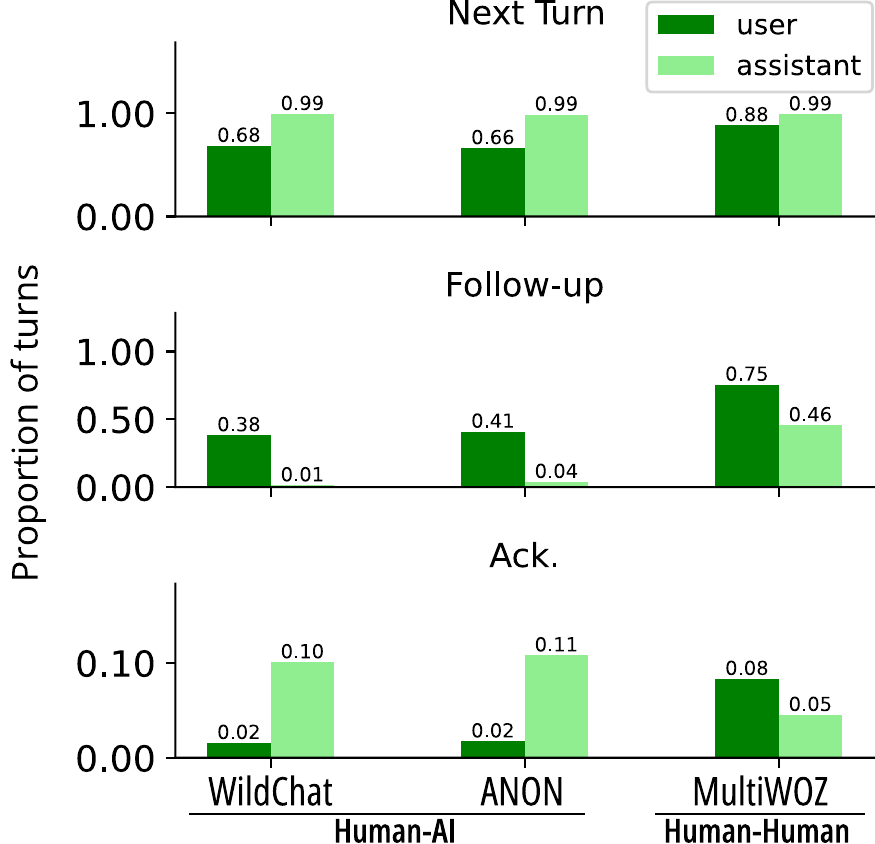}
    \caption{\textbf{Advancing Acts.} On all datasets, assistants overwhelmingly initiate the next turn in conversation, given instruction-following tendencies (row 1). Despite this, users construct more follow-ups when initiating the next turn compared to LLM assistants (row 2). In contrast, on MultiWOZ, containing interactions between human users and human assistants, both generated follow-up questions. Finally, LLMs over-generate acknowledgment acts (row 3), offering a false sense of ``understanding.'' This is especially surprising, considering that humans repair and clarify more. 
    }
    \label{fig:pos_grounding}
\end{figure}

\paragraph{Results.}
\label{sec:classification_results}
We observe significant differences between people and LLMs when initiating actions aimed at grounding. In datasets where LLMs serve as assistants, we observe that grounding acts are taken primarily by people. People \emph{repair and reformulate} instructions at high rates; averaged across human-AI interaction datasets, \hlgraytab{5\%} and \hlgraytab{18\%} of human turns are labeled as reformulate and repair respectively (Figure ~\ref{fig:neg_grounding}). In contrast to human-LLM interactions, human-human interaction data (MultiWOZ) has fewer reformulate (\hlgraytab{4\%}) and repair (\hlgraytab{3\%}) acts initiated by humans. 

Session \emph{restarts} serve as a final fallback when repair or reformulation fail to address a communicated failure. We focus on users with multiple interactions on WildChat (the only dataset that includes user IDs) and identify if a session begins with a repair or reformulation of an earlier instruction issued within the last 30 minutes. \hlgraytab{10.7\%} of sessions are restarts of a session in the last 30 minutes---exceeding the rate of repair in a conversation.

Before a human ends up addressing grounding, an ideal LLM assistant would have proactively engaged in clarification. However, we find that LLMs clarify at \textit{significantly} lower rates ($p < 0.01$, t-test) compared to humans repairing. In fact, the opposite occurs: people clarify LLM outputs (\hlgraytab{6\%}) 3 times as much as LLMs clarify user instructions (\hlgraytab{2\%}; Fig. \ref{fig:ambig_grounding}). In contrast, humans and wizard-of-oz-ed assistants clarify at similar rates (MultiWOZ; \hlgraytab{3\%} human user versus \hlgraytab{2\%} human assistant). 

Beyond repairing/reformulating, people regularly ask \emph{directed follow-up questions} when signaling at successful grounding (Figure ~\ref{fig:pos_grounding}). Across WildChat and \copilot{}, users ask \hlgraytab{15.6} times more follow-ups than LLM assistants. This disparity is less apparent in human-human interaction data, where human users only follow up \hlgraytab{1.7} times more. Instead of generating follow-ups, LLM assistants regularly over-respond (\hlgraytab{45\%} of assistant turns), generating verbose responses that answer more than what the user asked for. Humans rarely overrespond---both when interacting with LLMs (\hlgraytab{0\%}) or when roleplaying an assistant (\hlgraytab{5\%}).

\begin{figure}
    \centering
    \includegraphics[width=\columnwidth]{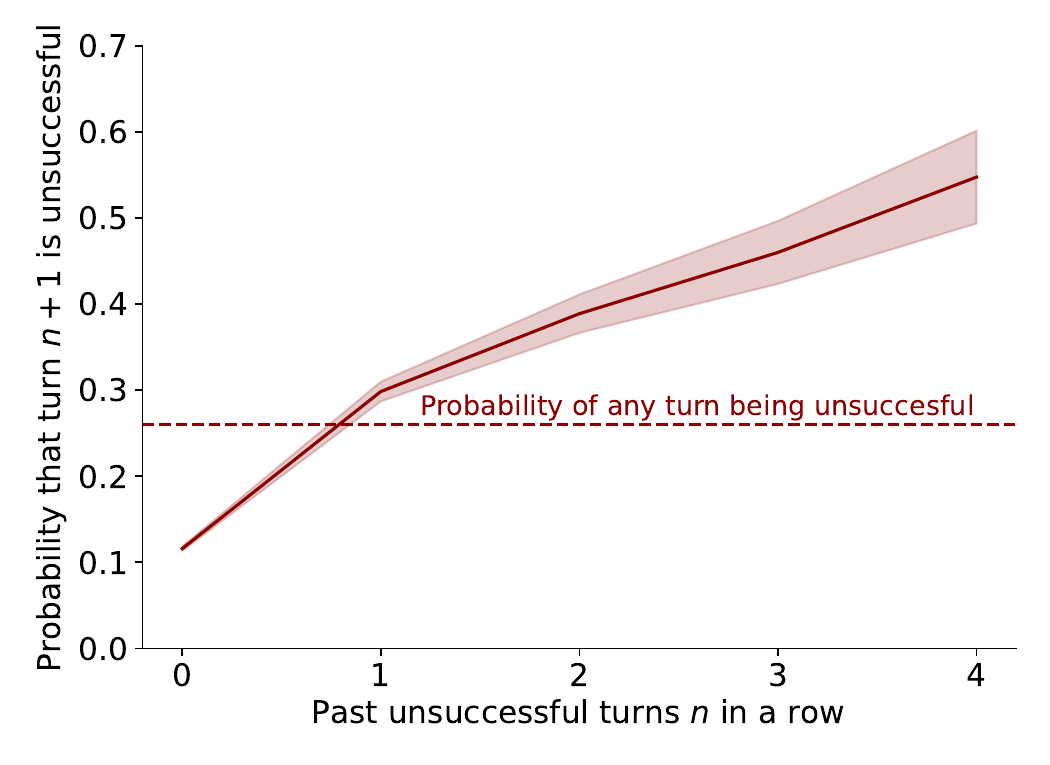}
    \caption{\textbf{Compounding Grounding Failures.}
    The plot shows the probability that the next turn in WildChat will contain an \hlsecondarytab{addressing} act given that the previous $n$ turns were addressing for $n \in [4]$. If a user signals addressing in an interaction, we observe that the following turns are more likely to be addressing. For example, the conditional probability rapidly increases from $P(m_0\in \mathcal{U}) = 0.12$ to $P(m_1\in \mathcal{U} |m_0\in \mathcal{U}) = 0.30$.}
    \label{fig:compounding}
\end{figure}

Early grounding patterns also have \emph{rippling effects}: we find evidence of compounding advancing \textit{and} addressing patterns as a conversation progresses (Fig. \ref{fig:compounding}). Let $m_i$ denote the message at turn $i$, which can be an advancing grounding act $\{m_i\in \mathcal{S}\}$ or addressing $\{m_i\in \mathcal{U}\}$ where $\mathcal{S}$ and $\mathcal{U}$ denote the sets of advancing and addressing grounding acts. The likelihood of an utterance representing an addressing act appearing in the first turn $P(m_0\in \mathcal{U}) = 0.12$ on WildChat, and $0.08$ on \copilot{}. We observe a compounding of addressing grounding in human-assistant conversations. If we assume that the turn before was addressing, we find that the likelihood of \textit{another} failed interaction triples: $P(m_1\in \mathcal{U} \mid m_0\in \mathcal{U}) = 0.30$ for WildChat and $0.32$ for \copilot{}. A similar effect appears for advancing acts: early signs of effective conversational grounding similarly snowballs, with $P(m_0\in \mathcal{S}) = 0.32$ $\rightarrow$ $P(m_1\in \mathcal{S} \mid m_0\in \mathcal{S}) = 0.47$ on WildChat and $0.23 \rightarrow 0.44$ on \copilot{}.

\subsection{Forecasting Users' Grounding Acts}
\label{sec:forecasting-grounding}
In light of the compounding effect of grounding failures (Fig. \ref{fig:compounding}), we pursue the possibility of lowering the probability of grounding failures byintervening \textit{before} a failure occurs. So far, we have introduced a GPT-based annotator (from \S\ref{sec:modeling_classification}) that annotates conversations with grounding acts in a post-hoc fashion. However, our prompted annotator only identifies opportunities for grounding \textit{after} they happen. Identifying or curating tasks where grounding failures emerge requires predicting the likelihood of future grounding patterns. To this end, we train a forecaster that predicts grounding acts in the next turn. We focus on WildChat as it is the only publicly availible human-LLM interaction dataset in our analysis.

\begin{table*}[t]
    \centering
    \small
    \begin{tabular}{l|ll}
        \toprule
        Category & Prompt from \textsc{Rifts} Benchmark \\ 
        \midrule
        Advancing & Write a Main heading about a brand name FFF Digital , which is a digital marketing agency \\
        % \cmidrule{2-4}
        & Suggest a name for a technical blog consisting of five characters at most, which is compatible with SEO  \\ 
        & 1 week out from my powerlifting meet and i'm not prepared [...] what should i do? \texttt{[omitted context]} \\ 
        \midrule
        Addressing  & Blackburn rovers vs West Bromwich albion prediction \\ 
        & I need to remove a heart \\
        & \texttt{[snippets of code with no prompt]} \\
        % & are u up to date \\
         % & \omar{TODO} \\
        \midrule
                 Disambiguating & What causes tailbone pain? \\
         & My friend not want to help me, what to [do] with him? \\
         & What happens when someone quits a job without having another one lined up? \\
        \midrule
         No Grounding & convert rust String to clap::builder::Str \\
         & Generate a full harvard references section for the following report: \texttt{[REPORT]} \\
         & Join now, Supplier! or  Supplier, Join us!   which one is better? \\
         % & \omar{TODO} \\
        \bottomrule
    \end{tabular}

    \caption{\textbf{Examples in \benchmark{}} fall into four categories. \emph{Advance} tasks are collaborative (e.g., resume building, diet planning, writing), where following up is necessary. \emph{Address} tasks are severely underspecified (e.g., contextless code snippets), or require capabilities LLMs lack (realtime information access)---these tasks need substantial clarification before any meaningful response is possible. \emph{Disambiguating} tasks are less severe, but still need context clarification (e.g., medical queries, relationship advice) for an ideal response. \emph{None} tasks are well-specified and factual, requiring no intervention. See \S\ref{appdx:cluster_analysis} for a lexical analysis of tasks in \benchmark{}.}

    \label{tab:example_tasks}
\end{table*}

\paragraph{Method.} Following a user message $m_i$, our goal is to forecast grounding act $g_{i + 1}$ associated with the next user message $m_{i + 1}$. We achieve this by repurposing \textit{conditional training}. Concretely, we append each user message $m_i$ with a forecasting grounding token $g_{i +1}$ if the \textit{future} turn $m_{i + 1}$ contains a grounding act. Consider the hypothetical training example below:

\vspace{10px}
\noindent \resizebox{\linewidth}{!}{
    $\begin{array}{l}
        \underbrace{\texttt{User: Help me write this section}}_{m_0} \hspace{5px}\underbrace{\hlsecondarytab{\texttt{addressing}}}_{g_1} \\
        \underbrace{\texttt{Assistant: Sure, here's the section...}}_{r_0} \\
        \underbrace{\texttt{User: \hlsecondarytab{Wrong section. I meant 5.2.}}}_{m_1}
    \end{array}$
}
\vspace{1px}

\noindent In the example above, we append a new forecasting token $g_1 = \hlsecondarytab{\texttt{addressing}}$ after the user's initial message $m_0$, since the user's following turn is a repair. We use our validated grounding acts labeler (\S\ref{sec:classification_results}) to obtain all $g_i$, using the high-level grounding categories as labels (e.g. \hlprimarytab{advancing}, \hlgraytab{disambiguating}, and \hlsecondarytab{addressing}). We finetune Llama-3.1-8B on sequences of form 
\begin{tikzpicture}[baseline=(current bounding box.center), node distance=0pt]
\node (langle) {\texttt{<}};
\node (mi) [right=-4pt of langle] {\texttt{m$_{i}$}};
\node (comma1) [right=-4pt of mi] {,};
\node (gi1) [right=-4pt of comma1] {\texttt{g$_{i+1}$}};
\node (comma2) [right=-4pt of gi1] {,};
\node (ri) [right=-4pt of comma2] {\texttt{r$_{i}$}};
\node (comma3) [right=-4pt of ri] {,};
\node (mi1) [right=-4pt of comma3] {\texttt{m$_{i+1}$}};
\node (comma4) [right=-4pt of mi1] {,};
\node (gi2) [right=-4pt of comma4] {\texttt{g$_{i+2}$}};
\node (comma5) [right=-4pt of gi2] {,};
\node (ri1) [right=-4pt of comma5] {\texttt{r$_{i+1}$}};
\node (comma6) [right=-4pt of ri1] {,};
\node (mi2) [right=-4pt of comma6] {\texttt{m$_{i+2}$}};
\node (dots) [right=-4pt of mi2] {$\ldots$};
\node (rangle) [right=-4pt of dots] {\texttt{>}};

\draw[->] (gi1.north) -- ++(0,0.10cm) -- ($(mi1.north)+(0,0.10cm)$) -- (mi1.north);
\draw[->] (gi2.north) -- ++(0,0.10cm) -- ($(mi2.north)+(0,0.10cm)$) -- (mi2.north);

\end{tikzpicture} where $r_i$ is the assistant response. 

At inference time, we can provide any user message $m_i$ and analyze the predicted likelihoods (i.e. logits) of our grounding tokens $g_{i + 1} \sim P(\cdot | m_i)$ that are generated right after. For example:

\vspace{8px}
\noindent \resizebox{0.8\linewidth}{!}{
    $\begin{array}{l}
        \underbrace{\texttt{User: Help me with this}}_{\mathrm{\texttt{prompt}} \hspace{3px} m_0} \hspace{5px}\underbrace{\hlsecondarytab{\texttt{addressing}}}_{\mathrm{\texttt{completion}} \hspace{3px} g_1}
    \end{array}$
}
\vspace{5px}

This enables us to predict---from just a user query alone---the user's predicted grounding outcome \emph{independent of the model's response!} This is an especially challenging learning problem: we effectively marginalize over all possible assistant responses. Successfully learning this model enables early intervention. All hyperparameters used in the training process are outlined in Appendix~\ref{appdx:forecaster_training}. Beyond our finetuned forecaster, we also evaluate as a baseline few-shot prompted GPT-4o-mini (details in Appendix \ref{appdx:forecaster_prompt}).

\paragraph{Results.} We find that our GPT-4o-mini few-shot baseline performs near-random, with Macro ROC AUC = 0.51. This result is not very surprising as forecasting is a challenging task: we must predict grounding acts $g_{i + 1}$ without directly observing assistant responses $r_i$. Our intuition is that it’s much easier to look at an entire conversation and label the conversation for grounding acts post-hoc than it is to forecast a likely grounding act without being able to ``see'' future turns. Our finetuned forecaster, however, performs significantly better (0.61).  Full experimental results are in Appendix~\ref{appdx:forecaster_training}. In the next section, we draw representative samples from the forecaster, constructing a benchmark where users are (un)likely to initiate a grounding process.

%% file: sections/06_benchmark.tex
\section{\benchmark{}: A Grounding Benchmark}
\label{sec:final-eval}

We showed that LLMs fail to generate grounding acts in two settings. They rarely:
\begin{itemize}
    \item \textit{Clarify} goals to reduce the rate at which a \emph{user is likely to address grounding}.
    \item \emph{Follow-up} to advance grounding, instead of relying on the \emph{user to take initiative.}
\end{itemize} 
To characterize these behaviors across multiple LLMs and evaluate interventions, we introduce a new benchmark, \benchmark. \benchmark{} consists of $\approx 1.8K$ tasks designed to test if LLMs can generate grounding acts when needed and withhold appropriately.

\paragraph{Dataset Details.} \benchmark{} consists of a final combined set of 1740 tasks (split counts in Appdx. Table \ref{tab:rift_splits}). Each task in \benchmark{} is a prompt drawn from an initial instruction in WildChat. Tasks are stratified based on how the user is predicted to continue the conversation: namely, are users predicted to \hlprimarytab{advance} grounding, \hlsecondarytab{address} a failure, or \hlgraytab{disambiguate} following an LLMs response (examples in Table \ref{tab:example_tasks}). For these tasks, we would expect an LLM to take initiative---clarifying or following up appropriately. In addition, we include tasks where the user is expected to do none of the above, perhaps by switching the topic or ending the interaction altogether. Here, no grounding is required.

\paragraph{Curation Process.} We construct \benchmark{} by filtering prompts from WildChat, using the predicted grounding act of the forecaster from \S\ref{sec:forecasting-grounding}. Forecaster predictions inform us on whether a \emph{clarification} or \emph{followup} action will be required in the conversation. In building \benchmark{}, we implicitly hypothesize that for some prompts $m_{i}$, regardless of what the LLM replies with ($r_{i}$), the user is in grounding trouble. In other words, if a user gives a query to a model that’s so severely underspecified (e.g. $m_{i}$ = “write me a resume”), it does not matter what any LLM responds with. The user must go back and forth to build common ground, since they never gave enough information in the initial prompt. \benchmark{} identifies this class of prompts, using the forecaster as a proxy.

We first filtered correctly predicted tasks from forecasters trained\footnote{\textit{Why train on each split?} To ensure fair evaluation, we create separate forecasters for train / val / test. This prevents leakage and enables researchers to develop interventions using the train forecaster before evaluating with the test forecaster.} on each split (train / val / test). For each grounding category, we then extracted the top 150 tasks with the highest likelihood of generating advancing, addressing, or ambiguous forecasting tokens. In other words, we repurpose our forecaster to curate representative tasks across each grounding act, sorting by the logit associated with each forecasting token. In addition, we sample 150 tasks that have a high likelihood of generating \emph{no} forecasting token, capturing tasks that do not require initiative. Finally, we apply basic quality controls (see Appendix \ref{appdx:filtering_criterion}).

\paragraph{Evaluating LLMs.}

\benchmark{} simplifies evaluation for any assistant model $P_\mathrm{assistant}$. Consider the two failure modes where LLMs do not take initiative: \textit{clarify} and \textit{followup}. Given a task from \benchmark{} in the \hlprimarytab{advancing} category, we would prefer $P_\mathrm{assistant}$ to proactively generate a followup. On the other hand, for tasks in \hlsecondarytab{addressing} or \hlgraytab{disambiguating}, we would expect $P_\mathrm{assistant}$ to generate clarification questions. In instances where we forecast no act from the user, we do not wish to see the model inefficiently engage in grounding activity.

To evaluate performance, we take an initial instruction $u_0$ from \benchmark{}, and sample the next turn $r_0$ from $P_{\mathrm{assistant}}(u_0)$. We then label $r_0$ with our validated grounding acts annotator (\S \ref{sec:modeling_classification}). To benchmark $P_{\mathrm{assistant}}$, we evaluate if the generated response $r_0$ clarifies/follows-up when appropriate. Concretely, we instantiate our two failure modes (\textit{clarify, followup}) in the following EVAL($u_0, r_0$) function and report an overall accuracy score:
\[
\begin{cases}
    1_{r_0 = \text{\textit{follow-up}}} & \text{if } u_0 \in \text{Advancing} \\
    1_{r_0 = \text{\textit{clarify}}} & \text{if } u_0 \in \text{Addressing} \cup \text{disambig.} \\
    1_{r_0 = \text{\textit{neither}}} & \text{if } u_0 \notin (\text{Addressing} \cup \text{dismbig.})
\end{cases}
\]

\begin{table}[t]
    \centering
    \small
    \resizebox{\linewidth}{!}{
    \begin{tabular}{ll|r}
        \toprule
        Model & Variant & \benchmark{} Accuracy \\
        
        \midrule
        \textbf{GPT} & 4o & 25.26 $\pm$ 3.54\\
        \phantom{\textbf{GPT}} & 4o-mini & 24.48 $\pm$ 3.51 \\
        \phantom{\textbf{GPT}} & o3-mini & 25.26 $\pm$ 3.54 \\
        \midrule
        \textbf{Claude} &  Sonnet 3.5 & 26.95 $\pm$ 3.57 \\ \phantom{\textbf{Claude}} &Opus 3 & 24.57 $\pm$ 3.51 \\        
       
        \midrule
        \textbf{Llama 3.1} & 8B Instruct & 24.22 $\pm$ 3.49 \\
        \phantom{\textbf{Llama 3.1}} & 70B Instruct & 23.88 $\pm$ 3.47\\
        \midrule
        \textbf{Llama 3.1} & 8B + \textsc{GROUND}& \textbf{54.48 $\pm$ 2.45}\\
        \bottomrule
        
    \end{tabular}
    }
    \caption{\textbf{Evaluating LLM grounding ability on \benchmark{}.} Frontier LLMs are ill-suited for grounding with humans on real-world tasks, with low accuracies across the board. A simple intervention (\textsc{+ GROUND}), based on our forecasters, can significantly improve LLM grounding ($\pm$ indicates a 95\% conf. interval).}
    \label{tab:rifts_results}
\end{table}

\paragraph{Off-the-shelf models struggle.} We evaluate a handful of open- and closed-source models on \benchmark{}' test set: OpenAI's GPT-4(o) series, Anthropic's Claude Sonnet 3.5 / Opus 3, and Llama 3.1 8 / 70B (Table~\ref{tab:rifts_results}). We find that \emph{all} off-the-shelf instruction-following models (avg. 23.23\% acc.) perform worse than random (33\%). All of our evaluated LLMs perform near perfectly for tasks that require no grounding initiative (No Grounding category, 96.09\%); this is unsurprising given instruction-following. However, LLMs fail to take appropriate initiative for any of the remaining categories (2.22\% of the time, Table \ref{tab:per_label_accuracies}). Reasoning-tuned models don't help either: o3-mini regularly begins reasoning without verifying grounding. 

\paragraph{A simple intervention.} To improve grounding capabilities, we turn again to our forecasters (\S\ref{sec:forecasting-grounding}). Depending on the train forecaster's prediction, we can selectively add a prompt (\textsc{+ GROUND}) that instructs the LLM to ask follow-up questions \emph{or} request clarification (prompts in Appendix \ref{appdx:intervention_prompt}). Concretely, we append a clarification prompt if our forecaster predicts \hlsecondarytab{address} or \hlgraytab{disambiguate}; or a follow-up prompt if our forecaster predicts \hlprimarytab{advance}. With this intervention, Llama 3.1 8B outperforms all other models by at least 32\%. Still, our intervention is far from perfect. \benchmark{} opens avenues for benchmarking new interventions, enabling easy evaluation of grounding capabilities in future work.

%% file: sections/07_discussion_conclusion.tex
\section{Discussion and Conclusion}
We characterized (\S\ref{sec:grounding_acts}) and measured (\S\ref{sec:data}-\ref{sec:modeling_all}) inadequate grounding in human-LLM interaction; and proposed a benchmark~(\S\ref{sec:final-eval}) to assess this gap. Several directions emerge:

\paragraph{Should we expect grounding behavior from LLMs?} Perhaps we should not be surprised that LLMs are unable to initiate grounding, defaulting instead to instruction-following. Models that are not trained to follow instructions are already biased towards instruction following behavior, likely because of the large presence of instruction following articles in pre-training mixes~\cite{hewitt2024instruction}. In addition, limitations in theory of mind and other metacognitive challenges may restrict the ability of models to engage in grounding interactions~\cite{sap2022neural, ullman2023large}. Training methods must counteract these limitations and biases. Still, we see promise in future methods that elicit grounding capabilities from LLMs; and \benchmark{} can serve as a resource to test these methods.

\paragraph{Towards LLMs that initiate grounding.} Decision-theoretic methods could guide when and how LLMs initiate grounding actions, based on inferred uncertainties in mutual understanding (see \citet{horvitz1999principles,mozannarAAAI2024}). Instruction tuning could be revised to incorporate grounding, and our forecaster could serve as a reward model in RLHF~\cite{ouyang2022training}. System prompts and dialogue management show promise, including prompts to disambiguate user intentions~\cite{chen-etal-2023-controllable}.

\paragraph{Benchmarking human-LLM grounding.} Building models that ground effectively with humans across a range of tasks requires effective benchmarks. \benchmark{} supports comparative analyses, enabling discussion on grounding competencies of new LLMs and interventions.

\section*{Limitations}

We considered grounding and engagement with \copilot{} in the absence of access to existing system meta prompts. System prompts can greatly shape the responses and provide specific guidance on the flow of dialogue. \benchmark{} was also collected by filtering WildChat using our forecaster; therefore, \benchmark{} will only reflect tasks seen in WildChat. In addition, tasks in \benchmark{} also depend on the LLMs used to serve WildChat (e.g. OpenAI LLMs). More specifically, our forecasters implicitly learn what tasks fail for the GPT models deployed in WildChat. Regardless, we observe that our final \benchmark{} tasks are challenging for all evaluated models. Finally, our annotator relies on GPT-4o-mini to label logs with grounding acts. While we did show that the annotator generally agrees with human judgment on a subset of the data (\S\ref{sec:classification_results}), the annotator is not perfect.

\section*{Ethics Statement}
Enhancing an LLM's ability to generate grounding acts (by initiating clarification and follow-up actions) raises potential concerns around privacy, as these actions may lead users to disclose sensitive information unintentionally. Balancing the need for grounding with the careful collection of only relevant information remains a significant challenge and an area for future research. Moreover, while effective grounding can improve interaction quality, it can also be misused in harmful contexts. Although our work focuses on improving grounding for constructive purposes, such as assisting users, these techniques could be exploited for harmful ends (e.g., manipulation, persuasion, or coercion in sensitive areas like political targeting). 

Finally, our description of human-LLM grounding \textit{does not imply that LLMs possess genuine understanding.} Like prior work, we use grounding acts to describe interaction processes that help align human expectations with LLM-generated responses~\cite{paek2003utility, brennan2014grounding, shaikh2024grounding}. While human conversation involves active mutual comprehension, the same cannot be said of LLMs. The use of grounding terminology in this work is intended as a conceptual tool to analyze how LLMs facilitate or hinder effective communication, not as an anthropomorphic assertion that they share human-like cognitive capacities.

\section*{Acknowledgments}
We appreciate the feedback from members of the Microsoft HAX team, the Stanford SALT Lab, Will Epperson, Michael Li, Jan-Philipp Fränken, Michael Ryan, Yanzhe Zhang, Qinan Yu, Shardul Sapkota and Ryan Li. 

%% file: sections/99_appendix.tex
\clearpage
\appendix
\onecolumn

\section{Dataset Splits and Details}

\subsection{Full Datasets}
\label{appdx:full_dataset_splits}

Table~\ref{tab:full_dataset_counts} outlines the number of dialogues and messages in each of our final filtered datasets. 

\begin{table}[h]
    \centering
    \begin{tabular}{l|rr}
        \toprule
        Dataset  & Number of Messages & Number of Dialogues \\
        \midrule
        WildChat  & 110688 & 55344 \\
        \copilot{} & 26200 & 13100\\
        MultiWOZ & 980 & 490\\
        \bottomrule
    \end{tabular}
    \caption{Number of messages and dialogues across our three analyzed datasets.}
    \label{tab:full_dataset_counts}
\end{table}

\subsection{Descriptive Analysis}
\label{appdx:cluster_analysis}
To quantify lexical differences between tasks in \benchmark{} and general instructions, we fit a Fightin' Words model between each cataegory and the full corpora~\cite{monroe2008fightin}. Fightin' Words reveals words that are associated with each particular text distribution, producing a log-odds ratio and a corresponding z-score. We select a sample of significant words that characterize each category in Table \ref{tab:fightin_words}. 

\begin{table}[ht]
\small
\centering
\begin{tabular}{l|l}
\toprule
\textbf{Label} & \textbf{Words (z-scores)} \\
\midrule
No Action & events (6.07), params (4.84), worksheets (3.78), quotation (3.42), answers (2.11), exam (2.03) \\
\midrule
Advancing & stock (5.24), dividend (3.77), regression (3.60), parents (3.23), investment (2.97), podcast (2.93) \\
\midrule
Addressing & sort (7.76), point (7.51), https (5.80), var (4.75), scan (3.45), merge (3.25), array (3.16) \\
\midrule
Disambiguating & bitcoin (4.66), chatbot (4.24), cryptocurrency (4.01), unauthorized (2.39), beginners (2.19), friends (2.02) \\
\bottomrule
\end{tabular}
\caption{Lexical cues from \benchmark{} reveal distinct task characteristics. 
\textbf{No Action} tasks involve users who are often trying to simply get answers for homework questions (e.g., worksheets), needing no follow-up. 
\textbf{Advancing} tasks (e.g., "stock," "dividend") imply iterative interaction, as in investment management. 
\textbf{Addressing} tasks feature technical, underspecified language (e.g., "sort," "array") that requires extra context---users often submit code with no explicit task. Finally,
\textbf{Disambiguating} tasks (e.g., "bitcoin") indicate a need for clarification on topics with inherent uncertainty: bitcoin, for example, is volatile, and beginners often have to clarify when learning.}
\label{tab:fightin_words}
\end{table}

\section{Training Forecasters}
Forecasting grounding is a challenging task; we are trying to predict if a user will have trouble for a task \textit{without} observing an LLMs generation. In other words, we can only use the task to predict future grounding patterns. We trained all models for 5 epochs (picked using the validation set), with learning rate 5e-5, batch size of 1, and 16 gradient accumulation steps (e.g. effective batch size of 16). All training occurred on an H100 80GB GPU. Below, we outline training optimizations that helped improve forecasting performance. 

\label{appdx:forecaster_training}
\subsection{Subsample Grounding Acts}
We subsampled WildChat data to include equal amounts of each forecasted grounding act before training. Inequal splits would result in the forecaster always predicting the majority class. To build our train/val/test splits, we sampled the maximum number of tasks (1630) possible from each of our forecaster categories ($1640\times 4$ for \texttt{fix, followup, continue, end}) while ensuring that each task was equally represented.

\subsection{Don't Mask User Tokens}
In addition to adding forecasting tokens, we make a modification to standard LLM finetuning/inference practices: we do not mask user utterances in the loss, training on user input. While the general effects of masking user tokens are mixed~\citep{huertaenochian2024instructionfinetuningdoesprompt, shi2024instruction, gottesman2024mask}, our setting requires the modeling of user input, as we seek to understand and assist with user grounding.  Because we do not mask user tokens, we can additionally simulate user inputs with past interaction data.

\subsection{Reweight Control Tokens At Train Time}
We seek to encourage our model to learn our added forecasting tokens alongside the language modeling objective. The standard MLE objective optimizes a model's parameters $\theta$ with respect to a sequence $x$: $\mathcal{L}(\theta) = - \sum_{t=1}^{T}\log p_\theta(x_t | x_{<t})$. However, a subsequence $x_{s...e}$ consists of forecasting tokens, which we want to emphasize---especially since these tokens did not undergo pretraining. At training time, we upweight these tokens by $\lambda = 2$. Our final loss is below:
\[
\mathcal{L}(\theta) = - \sum_{t=1}^{T} 
\begin{cases}
\lambda \cdot \log p_\theta(x_t | x_{<t}), & \text{if } s \leq t \leq e \\
\log p_\theta(x_t | x_{<t}), & \text{otherwise}
\end{cases}
\]

\subsection{Experimental Results}

\begin{table}[h]
    \centering
    \small
    \begin{tabular}{l|rr}
        \toprule
        & Few-shot GPT-4o-mini  & Llama 3.1 FT \\
        \midrule
        Followup & 0.52 & 0.61\\
        Fix & 0.49 & 0.60 \\
        Next Turn & 0.52 & 0.67 \\
        End & 0.51 & 0.57 \\
        \midrule
        Macro & 0.51 & 0.61 \\
        \bottomrule 
    \end{tabular}
    
    \caption{\textbf{Forecasting performance.} Per-label and Macro AUROC for forecasting task on the WildChat test set, conditioned on the initial prompt. }
    \label{tab:forecasting_perf}
\end{table}
In Table \ref{tab:forecasting_perf} we show the performance of our fine-tuned Llama 3.1 model compared to a few-show prompted GPT-4o-mini at forecasting grounding acts. We note the per-label and macro AUROC on the WildChat test set.

\section{Filtering Criterion}
\label{appdx:filtering_criterion}
While we sample tasks from the forecaster tails to construct \benchmark{}, we manually filter out tasks that ask for explicit content generation or ask the LLM for API keys, gift card codes, etc. Additionally, we passed tasks through the OpenAI moderation API, and filter out flagged tasks.

\section{\benchmark{}}

\begin{table}[htbp]
\centering
\small
\begin{tabular}{l|rrrr}
\midrule
\textbf{Model} & \textbf{Addressing (\%)} & \textbf{Advancing (\%)} & \textbf{No Grounding (\%)} & \textbf{Disambiguating (\%)} \\
\midrule
o3-mini & 4.14 & 1.35 & 90.65 & 8.22 \\
gpt-4o-mini & 2.07 & 0.68 & 96.40 & 2.07 \\
gpt-4o & 2.76 & 1.35 & 98.56 & 2.05 \\
claude-3-opus & 1.38 & 1.35 & 96.40 & 2.74 \\
claude-3-5-sonnet & 2.76 & 2.03 & 97.84 & 4.79 \\
Meta-Llama-3.1-8B & 0.69 & 2.03 & 96.40 & 1.37 \\
Meta-Llama-3.1-70B & 0.00 & 2.03 & 96.40 & 0.68 \\
\midrule
\textbf{Average} & 1.97 & 1.55 & 96.09 & 3.13 \\
\bottomrule
\end{tabular}
\caption{Per-label accuracies for various models on \benchmark{}. Most models correctly withhold initiation for tasks that require no grounding. However, this comes at a cost: models struggle at taking initiative for all other categories.} 
\label{tab:per_label_accuracies}
\end{table}

\begin{table}[t]
\centering
\small
\begin{tabular}{l|rrr}
\toprule
Category & Train & Val & Test \\
\midrule
Advancing  & 147 & 142 & 148\\
Addressing  & 144 & 143 & 145\\
Disambiguating & 146 & 147 & 146 \\
No Grounding & 146 & 147 & 139 \\
\bottomrule
\end{tabular}
\caption{\textbf{\benchmark{}} evaluation splits across 1740 total tasks.}
\label{tab:rift_splits}
\end{table}

\subsection{More Tasks}
\label{appdx:more_tasks}
We include a handful of tasks from Rifts directly in the paper (Table \ref{tab:example_tasks}). Our full benchmark and set of tasks can be found at our anonymous repository link, under the rifts folder: \href{https://anonymous.4open.science/r/rifts-B7E4/}{https://anonymous.4open.science/r/rifts-B7E4/}

\section{Prompted Models}
\label{appdx:prompts}

All of our prompts are located in the prompts folder in the following anonymous repository: \href{https://anonymous.4open.science/r/rifts-B7E4/}{https://anonymous.4open.science/r/rifts-B7E4/}. We detail each prompt used in our analysis below:

\subsection{Grounding Acts Labeling Prompt}
\label{appdx:grounding_acts_prompt}

We construct a few-shot prompt to annotate grounding acts across our datasets. The first author prompt-engineered a prompt on a small validation set. Our full prompt is availible in the \href{https://anonymous.4open.science/r/rifts-B7E4/}{anonymous repo}.

\begin{table}[]
    \centering
    \small
    % \resizebox{\linewidth}{!}{
    \begin{tabular}{l|rr}
        \toprule
        \textbf{Grounding Act} & \textbf{Support} & \textbf{4o-mini few-shot (F1 Score)} \\
        \midrule
        Next Turn     & 38      & 0.84 \\
        Acknowledge   & 4       & 0.67 \\
        Follow-up     & 17      & 0.80 \\
        Overcontinue  & 21      & 0.76 \\
        Clarify       & 7       & 0.67 \\
        Repair        & 10      & 0.74 \\
        Reformulate   & 11      & 0.78 \\
        \cmidrule{1-3}
        \textbf{Macro} &    108     & \textbf{0.75} \\
        \bottomrule
    \end{tabular}
    % }
    \caption{F-1 for Human-LLM grounding acts classification on a withheld test set of 30 conversations from our selected dialogue datasets. In the few-shot setting, \texttt{GPT-4o-mini} has fairly high F-1.}
    \label{tab:acc}
\end{table}

\subsection{GPT Forecaster Baseline Prompt}
\label{appdx:forecaster_prompt}

Alongside our finetuned Llama forecaster, we test a prompted baseline. We provide our prompted baseline with a few shot (task, future grounding act) pairs sampled from the forecaster train set. Our full prompt is in the \href{https://anonymous.4open.science/r/rifts-B7E4/}{anonymous repo} under the prompts folder.

\subsection{Intervention Prompt}
\label{appdx:intervention_prompt}

Our \textsc{Ground} intervention relies on two prompts, both in the \href{https://anonymous.4open.science/r/rifts-B7E4/}{anonymous repo} under the intervention folder. Specifically, we construct a follow-up prompt and a clarification prompt. Both prompts directly instruct the LLM to generate a clarification question or generate the answer + a followup. Our intervention is \textit{dumb by design}---the forecaster decides when to employ a static prompt. In instances where our train forecaster predicts \hlsecondarytab{address}, we enable the clarification prompt. Similarly, when our forecaster predicts \hlprimarytab{advance}, we employ the followup prompt. Given the improvements we see with our intervention, we expect that models trained to initiate grounding acts will substantially improve on \benchmark{}.

\section{License Information}

The Multiwoz \cite{eric2019multiwoz} dataset we analyze has the MIT license, the license file is available at \footnote{\url{https://github.com/budzianowski/multiwoz/blob/master/LICENSE}}.

The Wildchat \cite{zhao2024wildchat} dataset has the ODC-By license, license information is available here \footnote{\url{https://huggingface.co/datasets/allenai/WildChat/blob/main/LICENSE.md}}. By consequence \benchmark{} is also released under the ODC-By license.

We rely on the Llama 3.1 models, the license for those models is available here \footnote{\url{https://www.llama.com/llama3_1/license/}}.

We obtained permission from the ethics review board to analyze the \copilot{} data logs and release the analysis in this paper.